\pdfoutput=1

\def\year{2022}\relax
\documentclass[letterpaper]{article} 
\usepackage{aaai22}  
\usepackage{times}  
\usepackage{helvet}  
\usepackage{courier}  
\usepackage[hyphens]{url}  
\usepackage{graphicx} 
\usepackage{natbib}  
\usepackage{caption} 
\DeclareCaptionStyle{ruled}{labelfont=normalfont,labelsep=colon,strut=off} 
\usepackage{algorithm}
\usepackage{algorithmic}

\usepackage{microtype}

\usepackage{booktabs}   
\usepackage{subcaption} 
                        
\usepackage{amsmath}
\usepackage{booktabs}

\usepackage{multirow}

\usepackage[compact]{titlesec}

\usepackage{enumitem}

\setlist{nosep}
\setlist[itemize,1]{leftmargin=*, listparindent=0pt}
\setlist[description,1]{leftmargin=*, wide=0pt, labelsep=2pt}
\setlist[enumerate,1]{leftmargin=*}

\usepackage[colorlinks]{hyperref}
\usepackage[colorinlistoftodos, textsize=tiny]{todonotes}

\title{Human-Imitating Metrics for Training and Evaluating Privacy Preserving Emotion Recognition Models Using Sociolinguistic Knowledge}

\author{
    Mimansa Jaiswal, Emily Mower Provost
}
\affiliations{
  University of Michigan, Ann Arbor\\
    \{mimansa, emilykmp\}@umich.edu
}

\usepackage[switch]{lineno}  

\begin{document}
\maketitle

\begin{abstract}
Privacy preservation is a crucial component of any real-world application.  
But, in applications relying on machine learning backends, privacy is challenging because models often capture more than what the model was initially trained for, resulting in the potential leakage of sensitive information.  
In this paper, we propose an automatic and quantifiable metric that allows us to evaluate humans' perception of a model's ability to preserve privacy with respect to sensitive variables.  
In this paper, we focus on saliency-based explanations, explanations that highlight regions of the input text, to infer internal workings of a black box model. We use the degree with which differences in interpretation of general vs privacy preserving models correlate with sociolinguistic biases to inform metric design.  
We show how certain commonly-used methods that seek to preserve privacy do not align with human perception of privacy preservation leading to distrust about model's claims.
We demonstrate the versatility of our proposed metric by validating its utility for measuring cross corpus generalization for both privacy and emotion.
Finally, we conduct crowdsourcing experiments to evaluate the inclination of the evaluators to choose a particular model for a given purpose when model explanations are provided, and show a positive relationship with the proposed metric.
%
To the best of our knowledge, we take the first step in proposing automatic and quantifiable metrics that best align with human perception of model's ability for privacy preservation, allowing for cost-effective model development.
\end{abstract}

\section{Introduction}

Privacy has been an important concern in deploying black box machine learning algorithms in the real world. 
This concern stems from the ability of multi-parameter neural networks to memorize and replicate training data, and its impact on both the re-identification of samples~\cite{carlini2019secret} used for training the model and potential to base classification or generation decisions on these ``unintentionally'' learned sensitive variables.
Many web based applications claim are trained using de-identified user data and hence private to build implying trustworthiness.
While it might be true that that these models do not explicitly make use of any protected variables, they often end up learning demographic variables in the generated representations~\cite{sun2019mitigating}.
In this paper, we evaluate how metrics commonly used to evaluate a model's prediction accuracy and privacy preservation ability do not necessarily correlate to human trust and judgement of how private a model's working is.
We then propose a cost-effective human-imitating explainable metric that can be used to maximize the model's ability to preserve privacy as perceived by humans, without incurring the huge costs of human evaluations and maintaining performance on standardized metrics.

Previous studies have verified that representations learned from machine learning models trained to predict emotion also encode demographic factors~\cite{jaiswal2020privacy}. 
This can range from effects such as the model using gender encoding in the input representation for text auto-completion to sound a `particular' way.
But these models are also used to make decisions that can have serious consequences on a person's quality of life, such as the model using gender-based information in making hiring decisions~\cite{emohiring}.
Knowing that publicly deployed ML models can imbibe this hidden demographic information leads to humans' distrust in the model's ability to preserve privacy. 
This may lead to users be unwilling to adopt helpful applications such as passive mental health monitoring, for the fear of revealing their personal information~\cite{gerke2020ethical}.

To counter the problem of data leakage or privacy preservation, researchers have proposed methods spanning multiple fields such as in vision, language, and speech~\cite{bengio2013representation, wang2020privacy}. 
One common way to mitigate these concerns is to introduce 
privacy preserving mechanisms that avoid implicit encoding of protected variables in the generated representations, through either introduction of noise in the training data or intentionally training the model to not learn a private variable.
In the field of natural language processing, researchers have proposed methods such as debiasing, model re-training, model distillation2 to improve privacy preservation of the model for different conditions and required privacy protections~\cite{liu2021machine}. 
Most of these models though are evaluated on two major criterion, performance on the desired prediction task (primary task, which is emotion recognition, in our case), and, the performance of the model to be able to predict the private variable (in our case, gender), either from the perspective of the original model or an adversarial action. 
Hence, the most common metric used in these papers are usually f1-scores or accuracy values.
While these evaluation methods are good for non-human-facing prediction models (such as, dam water level prediction), they miss the mark in being able to quantify a user's perception of the model, both, for the primary and the privacy preservation task, which is a necessary component for the adoption of AI-based decision systems. But, a challenge remains: the major hindrance towards using human judgement as a metric has been non-quantifiable and expensive to obtain human evaluations.

In this paper, we introduce a metric, which we call the emotion-privacy (EP) metric  that will quantify human judgement of the performance of, and, preference for, any given model with respect to both emotion recognition and privacy preservation.  For the privacy preservation task, we focus on gender prediction, acknowledging that the benchmark emotion recognition datasets are restricted to binary gender labels, which may not align with how individuals self-identify. 
To do this, we consider two lists of word-tokens both of which are weighted.  The first list has word tokens that have been identified to be indicative of gender
from sociolinguistic literature.  The second list contains word tokens that the model learns to unlearn, again with associated weights, when being trained to preserve privacy.  We identify words that occur in both lists and, for each word, multiply the weight of that word in each list.  Next, we identify words that occur in only one list, multiplying that word's weight by  -1, indicating the mismatch.  This creates an importance value for each word.  Finally, for an input sample, we sum the importance values for each word in the text and divide by the total number of tokens.  However, the EP metric refers to the dataset as a whole.  We calculate the EP metric over the complete dataset by taking an average of this value over all samples in the dataset.

We then conduct both, empirical and crowdsourcing-based experiments to test the goodness of the proposed metric. 
We imbue the emotion recognition models with privacy preservation through multiple state-of-the-art strategies: adversarial training, bias-fine tuning, and dataset augmentation.
We first evaluate the performance of these trained models on three major criterion: generalizability, performance of emotion prediction, and performance of an adversary aiming to obtain private information.
We then conduct crowdsourcing experiments to find how these multiple privacy preservation techniques are perceived by humans, asking them questions related to trust in the model and preference of a model.


We find that our 
metric has consistently significant high correlation with user trust and preference, and hence can be used to train or evaluate privacy preservation models.
We remind that the computation of the proposed metric is independent of the private variable and can be applied as long as there exist sociolinguistic studies to draw from.  This suggests that this approach can be used for alternative variables of interest. 
Additionally, the computation of our metric is independent of the chosen method for model interpretation, and relies only on having a weighted word list, irrespective of the reason and motivation behind the obtained ordering.
We offer some suggestions on how to obtain these lists for other commonly studied protected variables, such as race and age.
\section{Related Work}

Model interpretability is a critical component of machine learning systems that are designed to interact with humans.  Models that are interpretable have the ability to explain \emph{why} they are making decisions, or in the context of privacy, \emph{why} users should have faith that they are preserving users' sensitive information.  We divide our related work section into three parts: 1) privacy preservation or debiasing, 2) interpretability methods, and 3) human trustworthiness.

\subsection{Privacy Preservation or Debiasing}
Researchers in fair algorithmic approaches have investigated approaches to develop models that are invariant to particular sensitive attributes, to obtain debiased word embedding~\cite{bolukbasi2016man}, ensure fairness parities~\cite{corbett2018measure}, and train debiased hate speech classification~\cite{davidson2019racial}.  These approaches often use the accuracy of the demographic variable prediction as the fairness metric.  
In this paper, we instead evaluate how interpretations can be used as an evaluation metric, one that measures the model's performance on the primary task and also how closely the rules learned by the machine learning system for the optimized variable (e.g., emotion, privacy, etc.)  are aligned with human trust. 

Researchers have investigated the impact of privacy preserving approaches by creating artificially augmented datasets that change the data distribution.  For example, Sun et al. created an augmented dataset that by changing the bias towards opposite gender in their original data set.
They then trained on the union of the original and data-swapped sets~\cite{sun2019mitigating} and found that the resultant models capture less racial information, and are more secure to adversarial effects used to obtain private information. 

Researchers have also investigated bias fine-tuning to reduce bias in machine learning models by initializing models with a large unbiased dataset with the aim of retaining that `unbiased' nature when fine-tuned~\cite{sun2019mitigating}. However,  work by~\cite{jaiswal2020privacy} shows how these models though can still introduce privacy violations while obtaining promising performance over privacy evaluation metrics. 

\subsection{Model Interpretation and Explanation}
New tools are being developed to create deep learning models that are more interpretable~\cite{kim2016examples, szegedy2013intriguing}. Interpretability has been used for auditing algorithms' predictions~\cite{belinkov2019analysis}. Recently, explanation methods have also been used as human aids, i.e., using summarization for medical records, such as Subjective, Objective, Assessment and Plan (SOAP) notes~\cite{krishna2020generating} or using explanations for model debugging~\cite{nushi2018towards}. 
Other methods, beyond input saliency, include multiple other approaches such as representative samples and influence functions, which is succinctly summarized by~\cite{belinkov2019analysis}.
While major interpretation methods cater to explaining models to a general audience, our work aligns more with use of interpretability to assess and debug a trained model by a model designer. Researchers have previously looked at using interpretability as a mechanism for auditing a model and checking the learned decision making flows~\cite{engler2021auditing}.

\subsection{Human Perception of ML Models}
To be deployed at a large scale, machine learning models must be robust and generalizable, while at the same time assuring users that the algorithms will appropriately handle their data.  One component of this goal is accomplished through the development of algorithms with strong privacy guarantees.  Another component is to create models that users will trust. Previous research has shown that humans are more likely to trust a model's decision if its explanation are aligned with how they, themselves, make predictions~\cite{sperrle2019human, ferrario2019ai, schmidt2019quantifying}. Further, it is easier for the user to trust a model and its capabilities if they are provided with information about how the model processes their data, and explanations from the model regarding how predictions were generated~\cite{schmidt2019quantifying}. 
This allows users to personally verify whether the model is implicitly making inferences tied to their sensitive demographic information (e.g., age or gender).  
Thus, explanations must be both plausible and faithful to the internal model workings. 
Researchers have previously investigated the utility of explanations for various stakeholders~\cite{preece2018stakeholders}, for inducing trustworthiness~\cite{heuer2020more} and have also explored the possibility of false privacy guarantees~\cite{pruthi2019learning}. 

In this work we look at how saliency methods, applied for model interpretability combined with sociolinguistic knowledge, can be used as a proxy metric for human perception of a model's ability for a given task.
This combination of model interpretation and expert knowledge allows us to have a cost-effective metric that not only captures the aspects of human judgement, but is also quantifiable and hence optimizable for machine learning algorithms.


\section{Research Questions}
\textbf{RQ1}: Is there a difference in saliency-based interpretations obtained from a model that is trained just for emotion recognition vs. another that is trained for emotion recognition while preserving privacy?
\\\textbf{RQ2}: Does the difference in the saliency-based interpretations, explanations that highlight regions of the input text and should thus match human judgment, actually align with human perception of goodness of privacy preservation?
\\\textbf{RQ3}: How can we quantify this difference to obtain a metric that is a proxy for human judgement of the goodness of privacy preservation - one that encodes both the model's ability to avoid private variable leakage and a person's belief that the model preserves privacy?
\\\textbf{RQ4}: Is the proposed metric versatile for quantifying human trust in other tasks beyond privacy preservation? We look at the primary task of emotion recognition, and investigate:
\\\textbf{RQ4a}: Does the proposed metric correlate with human's trust in the ability of a model to effectively recognize emotion?  
\textbf{RQ4b}: If yes, does the use of this metric, which has been designed to be both optimizable and human-aligned, in model training encourage generalizability? 

\section{Datasets}
We use three datasets for emotion recognition purposes, IEMOCAP~\cite{busso2008iemocap},  MSP-Improv~\cite{busso2016msp}, and MuSE~\cite{jaiswal2020muse}, focusing on only the text modality of each dataset.  
\textbf{IEMOCAP}, an audio-visual + motion capture dataset consisting of 10,039 utterances, was collected to understand how emotion expressions shape behavioral patterns.
The data consist of interactions between five mixed-gender pairs of actors (one male and one female, ten actors total). The data were labeled by 5 human annotators using emotion categories and dimensional attributes respectively. \textbf{MuSE}, an audio-visual dataset consisting of 2,648 utterances, was collected to understand the interplay between stress and emotion in natural spoken communication. The data were labeled by 232 human annotators, 3 per utterance using dimensional attributes. \textbf{MSP-Improv}, an audio-visual dataset consisting of 8,438 utterances, was collected to capture naturalistic emotions by 12 actors (6 male, 6 female). The data was annotated by ~5 evaluators per utterance for both emotion categories and dimensional attributes. 
We restrict our analysis to the data collected from the improvised turns (4,381 utterances). These data consist of fixed prompts, initiated first by one actor, and then the other.  This provides a controlled environment for the collection of emotional sentences, where the topic distribution is consistent across both genders, reducing the gender-bias.  This will allow us to train a model that is also less biased due to lexical content for the primary emotion recognition task.  The transcripts in both MuSE and IEMOCAP were generated by humans, the transcripts in MSP-Improv were generated using Microsoft Azure Speech-to-Text.
%
\textbf{IEMOCAP} is used to train the set of privacy preserving methods whose predictions and explanations are used in the crowdsourcing experiment.  \textbf{MSP-Improv} is used as an auxiliary dataset in bias-fine tuning, one of the privacy preservation methods.  \textbf{MuSE} is used to assess the generalizability of the models learned on the IEMOCAP or MSP-Improv+IEMOCAP (initialized on MSP, fine-tuned on IEMOCAP).  

\subsection{Labels}
The dimensional emotion labels are binned into 3 classes to represent \texttt{\{low, mid, high\}} for valence using binned averaged rating. The classes are defined as \{(low:[1,2.75]), (mid:(2.75,3.25]), (high:(3.25,5])\} for IEMOCAP and MSP-Improv and \{(low:[1,3.75]), (mid:(3.75,4.25]), (high:(4.25,9])\} for MuSE.
We treat the problem as a three-way classification problem, where the goal is to assign a label from \{\emph{low}, \emph{mid}, \emph{high}\} to a given utterance. 

\section{Methods}
\label{sec:privmethod}

\subsection{General Emotion Recognition Model~\emph{:General}}
We train all of the following models on IEMOCAP* (original set or modifications as mentioned below). For within dataset evaluation, we use the test dataset from IEMOCAP, whereas, for out of distribution data evaluation, we test the same model on MuSE dataset.
\label{sec:genmodel}
Our classification model is based on the base version of Bidirectional Encoder Representations from Transformers (BERT) model due to the prevalence of this approach~\cite{devlin2018bert}. We also use a pre-trained BeRT tokenizer for the model. We implement and fine-tune the model using the HuggingFace library~\cite{wolf2019huggingface}. 
We replace the pretraining head of the BERT model with a classification head, which is randomly initialized. We use the \emph{Trainer} module that takes in \emph{TrainingArguments} to then fine-tune the model for our purposes.
We use unweighted average recall (UAR) as our evaluation metric for the emotion classification to account for class imbalance~\cite{rosenberg2012classifying}. We train our models using PyTorch with RMSprop and a weighted cross-entropy loss function, and, use the validation set for early stopping after the results do not improve for five consecutive epochs. We run each experiment three times to account for random initialization of the parameters.

\subsection{Privacy by Adversarial Training~\emph{:PrivAdv}}
\label{sec:privadv}
We use an adversarial paradigm to train the models to preserve the privacy of the generated embeddings with respect to the demographic variable of gender. The main network is trained to unlearn gender using a Gradient Reversal Layer (GRL)~\cite{ganin2014unsupervised}. GRLs are a multi-task approach to train models that are invariant to specific properties~\cite{meng2018speaker}. We place the GRL function between the embedding sub-network and the gender classifier to obtain gender-invariant representations. 

\subsection{Privacy by Data Augmentation~\emph{:PrivAug}}
We create an augmented data set using gender-swapping to compare our proposed method and resultant metric to other successful approaches~\cite{iosifidis2018dealing}. We use a pronoun-based word list and create a gender-swapped equivalent for each sentence, for e.g., replacing ``he'' with ``she'', or ``his'' with ``hers'' and so on. Data augmentation does come with its own issues: double the training data size with no added primary task information, expensive list creation of gender based words in the dataset to be replaced, and nonsensical sentence creation~\cite{belinkov2019analysis}.

\subsection{Privacy by Bias Fine-Tuning~\emph{:PrivBias}}
Bias fine-tuning assumes the existence of an additional dataset that is unbiased with respect to the sensitive variable of interest, which is used to train an initial model.  The goal is that the resulting model should then no longer encode bias due solely to dataset characteristics.  We use the improvised turns subset of MSP-Improv for unbiased fine-tuning (the improvised prompts are unbiased with respect to gender). We then fine-tune this model on IEMOCAP. 

\subsection{Control Models}
We train and use the two models that follow to have maximally poor performance for privacy preservation.  This provides a quality control check for the crowdsourcing experiments, we can monitor annotation quality, and a negative training baseline for our proposed metric.
\textbf{\emph{GenderCtrl}: Explicitly training for gender classification} The GenderExp model is a multi-task model, which adds gender prediction on top of the general model. It is used as a control mechanism for any metric calculation, because it should ideally capture correlations that we are trying to avoid, leading to a low evaluation score on the chosen metrics (Section~\ref{sec:modelexp}).

\noindent \textbf{ArtNoise: Artificially Noisy Model} We create a parallel corpus of IEMOCAP to use as a control baseline. We add six artificially "noisy" features, i.e, \{'zq0', 'zq1', 'zq2', 'zx0', 'zx1', 'zx2'\}, such that they correlate specifically with both emotion and gender. For example, zq0 would be added to a random selection of male-low valence samples.  We then train a classification model to predict emotion on this dataset, and highlight salient features.
Because the added signals are completely correlated to various sub-classes in the dataset, it ensures that the model always learns these added tokens as salient features. Given these added tokens are meaningless, significant saliency of these tokens implies that the model explanations are neither good for judging privacy preservation nor for judging ability of emotion recognition~\cite{kaushik2019learning}.
This acts as a controlled mechanism, specifically an attention check, allowing us to discard any crowdsourced task where the crowdsourced worker selected the result from this model, because the words highlighted by this model are always nonsensical.

\section{Privacy Metrics via Human Perception}
\label{sec:iat}
People often rely on gender-based patterns when assessing emotion expression~\cite{brescoll2016leading}. The term, perceptual bias, is defined as the probability of any sample being specifically understood as a particular emotion based on the speaker's gender and not on how the information was expressed.
We aim to design a metric such that it can capture this bias to tease apart the linguistic indicators through which humans attribute gender to any written text~\cite{fosch2021little}. We focus on two types of sources to obtain a desired keyword list, (a) using word lists that have been tested using Implicit Association Test (IAT) to have population level bias between the genders and (b) sociological studies~\cite{newman2008gender}.
The IAT is designed to reveal attitudes and other automatic associations even for subjects who prefer not to express those attitudes. We use the gender bias IAT list provided by WEAT (Word Embedding Association Test)~\cite{swinger2019biases} to identify words that are known to have significant differential attitude towards a word.
Sociologists have discussed  that humans recognize gender in written communication using certain word categories~\cite{brescoll2016leading}. For example, women tend to use more hedge words but men are likely to use more referential language. We consider hedge words, tags, referential language, profanity, first person pronouns and politeness markers as the categories.
We use Linguistic Word Inquiry Count (LIWC) to find keywords or phrases that fall into the defined categories of biased perception indicators as mentioned in these studies.
 
\subsection{Sociolinguistically-Informed Wordlists}
While we focus on gender in this study (presently binary due to the manner in which gender was categorized in the target datasets), there are other variables that researchers have aimed to preserve privacy for, e.g., age, race etc. While we do not consider these variables in our study to the expansive scope, we collate some sources that can be used to inform development of these word lists for other variables.
Differences based on age in linguistic choices and perception have been extensively studied in which provides pointers to creating these lists, including words such as~\cite{eckert2017age}.
Similarly, race based linguistic differences have been documented in various population level, group level and case studies. These papers and books also provide a good source for keywords, such as~\cite{rickford2016raciolinguistics}, that can be used to create a `\emph{prior knowledge}' based list. 

\subsection{Model Explanations}
\label{sec:modelexp}
We use the Captum interpretability library for PyTorch~\cite{kokhlikyan2020captum}. Captum provides state-of-the-art algorithms to identify how the input features, hidden neurons, and layers contribute to a model’s output. We use the attribution algorithms implemented via integrated gradients~\cite{sundararajan2017axiomatic}. Integrated gradients represent the integral of gradients with respect to inputs along the path from a given baseline (absence of the cause) to input sample. The integral can be approximated using a Riemann Sum or Gauss Legendre quadrature rule. The output is a set of words for each model instance that contribute towards the prediction along with their attribution weights. We refer to these sets as attribution word sets.  

We will compare these sets from pairs of models and, in the rest of the paper, we will refer to differences between the sets as the ~\emph{model attribution difference}. It has two parts, the first is a weighted sum of the words the privacy preserving model avoids that are correlated to gender when compared to the general model, and, the second is a penalty for including words that are correlated to gender that weren't included in the general model (Section~\ref{sec:genmodel}).
We calculate the Emotion-Privacy (EP) metric,
which is an extension of general expectation overlap, which captures the overlap of salient words with the list of `expected' words, often used to measure agreement.
We extend our EP metric to not only capture the aforementioned overlap, but also be able to attribute change in overlap to the particularly introduced intervention (in this paper, privacy preservation) to provide meaningful evidence towards approaching causality.  The EP metric calculation is based on four components: (i) the number of samples in the model ($N$), (ii) the explanation set obtained from the generally trained model ($E(M_g)$) and the comparison model ($E(M_x)$), (iii) the number of words in the explanation set produced by the generally trained model ($G(M_g)$), and (iv) the above created list of implicitly gender-biased perception words ($L$) as:
\vspace{-1em}
\begin{equation}
\footnotesize
\begin{aligned}
&G(M_g) &=& \left \lvert \bigcup_{j\in L} \{\lvert w_j \rvert : w_j \in E(M_g)\} \right \rvert\\
&E_d &=& \left( E(M_g) - E(M_x)\right) \cap L \hfill\\
&E_{a} &=& \left( E(M_x) - E(M_g)\right) \cap L \hfill\\
&\text{EP} &=& {\frac {1}{N}}\sum_{i=1}^{N} \frac{\left(\sum_{j=1}^{\lvert E_d \rvert}\lvert w_j\rvert -\sum_{j=1}^{\lvert E_a \rvert}\lvert w_j\rvert\right)}{G(M_g)}
\end{aligned}
\end{equation}
\vspace{-0.4em}
To show the versatility of the metric, we also show a variant of the EP metric, as, emotion-generalization (EG), that captures the degree and direction of overlap of the salient words with the list of expected words.
We perform a linguistic analysis on the samples that the model gets right for the new dataset, vs. the ones it gets wrong, to investigate the difference in the generalizability of the model when either using or not using the EG metric. We use the National Research Council Canada - Valence, Arousal, and Dominance (NRC-VAD) Lexicon~\cite{mohammad2018obtaining} corpus consisting of ~20,000 unigrams that are annotated for the emotion axes. 

Conventionally, generalizability is assessed by training the model on one dataset and testing on another for the same task.
We calculate the probability that sample $i$ relies on spurious correlations ($Px_{i,sc}$) as the ratio of words in the machine explanation that are correlated in the wrong direction with the words in NRC-VAD over the total number of words. The probability of a model relying on spurious correlation (${{EG}_{not}}$) is the average of all sample probabilities. Consequentially, Emotion-Generalization ($EG$) is the estimated generalization capability of a model for emotion recognition.
\begin{equation}
\footnotesize
\begin{aligned}
&Px_{i, sc} = {\frac {1}{sN}}(c_n \left(\lvert w_j\rvert) \right)+c_p\left(\lvert w_j\rvert\right)\\
&{{EG}_{not}} = {\frac {1}{N}}\sum_{i=1}^{N} Px_{i, sc}\\
&EG= 1 - {{EG}_{not}}
\end{aligned}
\end{equation}
\noindent where $sN$ is the number of words in the sample, 
$c_n$ are words in the model explanation for a sample that are negatively attributed to prediction, and have the same binned value in NRC-VAD as ground truth label, $c_p$ are words in the model explanation for a sample that are positively attributed to prediction, and have the opposite binned value in NRC-VAD as ground truth label, $w_j$ is the weight of the attribution for word $j$ as obtained from CAPTUM integrated gradients, $N$ is the number of samples correctly predicted by the model only trained for the primary task of emotion recognition.


We compare the $EG$ for samples that the generally trained model correctly classifies.  This allows us to create a baseline system from which we can understand the increase in spurious correlations.
This will also allow us to understand how increases in spurious correlations lead to performance differences both within and across datasets.



\begin{table*}
\centering
\footnotesize
\caption{\footnotesize Results for valence (Val) prediction on IEMOCAP, \emph{I}, and MuSE, \emph{M}. The model names on the left refer to the models trained in their corresponding as described in Section~\ref{sec:privmethod}. $EP$ is the proposed metric to corrleate interpretation difference with curated lists (Section~\ref{sec:modelexp}). $EG$ is estimated generalization ability of the model. C(Em) and C(Pv) refer to the proportion of people who showed preference for a particular model for the task of emotion recognition and privacy preservation respectively. Random, Low and High EP refer to the sample categories as described in Section~\ref{sec:questions}
\textbf{\textit{ Bold}} shows significant improvement.Higher values for valence, $EP$ and $EG$ are better, and lower value for gender is better.. Significance is established using paired t-test, adjusted p-value$<0.05$.
}
\begin{tabular}{c|cccc|cc|cccccc}
\hline
&\multicolumn{2}{c}{IEMOCAP} & \multicolumn{2}{c}{MuSE} & \multicolumn{2}{c}{Proposed Metric} & \multicolumn{2}{c}{Random EP} & \multicolumn{2}{c}{High EP} & \multicolumn{2}{c}{Low EP}
 \\

                              & V-UAR          & G-UAR & V-UAR           & G-UAR & \emph{EP}  & \emph{$EG$} & \emph{C(Em)} & \emph{C(Pv)} & \emph{C(Em)} & \emph{C(Pv)} & \emph{C(Em)} & \emph{C(Pv)}
 \\\hline
General   & 0.648 & 0.711  & 0.577 & 0.742 &  - & 0.65 & 0.35 & 0.1 & 0.3 & 0.06 & \textbf{0.46} & \textit{0.28*}\\
PrivAdv   & \textbf{0.671} & 0.562 & 0.532 & \textbf{0.622} & 0.653 & 0.62 & 0.30 & 0.33 & \textbf{0.36} & \textbf{0.38} & 0.24 & \textit{0.24*} \\
PrivAug   & 0.636 & \textbf{0.551} & \textbf{0.591} & 0.643 & \textbf{0.680} & \textbf{0.68} & 0.26 & \textbf{0.39} & 0.31 & \textbf{0.43} & 0.18 & 0.21\\
PrivBias  & 0.620 & 0.621 & 0.528  &  0.688 & 0.563     &     0.60   &0.02 & 0.12 & 0.01 & 0.10 & 0.06 & 0.15 \\
GenderCtrl& 0.631 & 0.820 & 0.517  &  0.756 & -0.210      &    0.59    & 0.09 & 0.06 & 0.02 & 0.03 & 0.06 & 0.11\\
\hline
\end{tabular}
\label{tab:vallex}
\end{table*}

\section{Experimental Setup}
We analyze the performance of these models within and cross-dataset, and then from the perspective of user preference. We use Amazon Mechanical Turk (MTurk) for crowdsourcing annotations to understand how the combination of model explanations and predictions shape evaluators' trust as measured by their selection of a specific model for a given task amongst the choices mentioned in Section~\ref{sec:questions}.

\subsection{Model Construction}
We use IEMOCAP to train the emotion recognition models in a speaker independent training and testing paradigm.  We train six different models on the original train subset of the IEMOCAP dataset: \{PrivAdv: Privacy Preservation using Adversarial Training, PrivAug: Privacy Preservation using Data Augmentation, PrivBias: Privacy Preservation using Bias Fine-Tuning, GenderExp: Explicitly training for gender as a multi-task classification problem, Gen: Generally trained model for just emotion recognition, and, ArtNoise: Generally trained model on artificially introduced noisy data\}. 

\subsection{Model Generalizability}
Previous research has shown that multi-objective training on one dataset, with poor choices of the tasks, might lead to poorer domain adaptation, because the model tends to rely more on spurious correlations~\cite{sagawa2020investigation}.
Therefore, we look at how multiple privacy preservation methods, perform when tested on a different dataset, both for gender and emotion prediction. 
We train the model on the IEMOCAP training set and then test the model on the MuSE dataset. 

\subsection{Crowdsourcing Experiment}
\label{sec:questions}
We show the predictions and explanations for a subset of the IEMOCAP testing partition in a crowdsourcing experiment. We aim to select a representative sample population by considering various levels of valence, as well as, gender for a total of six bins. We select 20 random samples from each of the six classes to use as our baseline and then choose additional 30 random samples each from the set of samples that have a high EP score and those that have a low EP score as obtained in Section~\ref{sec:iat}, for a total 540 samples.  

We extract both explanations and predictions using Captum for each the 540 samples, using each of the five models (PrivAdv, PrivAug, PrivBias, GenderCtrl, ArtNoise, described above). The ArtNoise is used as a control baseline for human evaluation, ensuring that the evaluators are paying attention to the task at hand. We consider this option as an attention check and discard any response where the crowdsourced worker preferred the result and/or the explanation from this model (7.31\% samples were discarded in total).

We recruited annotators using MTurk from a population of workers with the following characteristics: 1) $>98$\% approval rating, 2) $>$ 500 approvals, 3) in the United States, and 4) native English speakers. Each Human Intelligence Task (HIT) was annotated by three workers. We ensured that all workers understood the meaning of valence using the popular qualification tests.
Each HIT took an average of one-minute. The compensation was $\$9.45/$hr. We present the predictions from these models and heat map based explanations to the workers and ask them to choose between these five models for each of these three questions: 1) Which model are you more likely to trust for emotion prediction? 2) Which model are you more likely to trust for privacy preservation? 

\section{Results and Discussion}

\subsection{RQ1: Quantifying Interpretation Differences}
\textbf{Hypothesis}: The differences in salient tokens obtained for general models vs those obtained for privacy preserving models should show avoidance of gender-related terms.
\\\textbf{Reason}: Previous researches have shown how inducing privacy preservation in a natural language avoids learning tokens indicative of gender~\cite{baron2020interpretable}.
\\\textbf{Result}: We show the shifts in the performance between models that are trained just for emotion recognition (General, Table~\ref{tab:vallex}) and those that are trained for emotion recognition and privacy preservation (adversarial training, PrivAdv, augmentation, PrivAug, bias fine-tuning, PrivBias, in Table~\ref{tab:vallex}). 

We first examine the change in the performance of the within-IEMOCAP valence classification task (column \emph{I} Valence UAR in Table~\ref{tab:vallex}, chance is 0.33).  We find that while there is a drop in valence prediction performance, moving from the General model to the privacy-preserving models, the difference lies in between $\pm$2\%. We also observe that while the General model can be used to recognize gender, the privacy-preserving models obstruct this information (column \emph{I} Gender UAR in Table~\ref{tab:vallex}, chance is 0.5).
We find that the top 65\% of the intersection between the EP and the implicit bias lists come from LIWC category-based words (for example: words implying hesitation, vagueness) and the other from IAT proposed lists (for e.g.: words implying self-confidence, arrogance) which indicates the usefulness of incorporating expert knowledge in model evaluation.

\subsection{RQ2: Crowdsourcing Validation of EP Metric}

\textbf{Hypothesis}: EP is significantly correlated with evaluators' perception of the ability of model for privacy preservation.
\\\textbf{Reason}: Previous researches have shown that humans perceive a model as more trustworthy and capable if its explanations align with how they would form their own judgments~\cite{schmidt2019quantifying}. Given that the metric is intentionally aimed to capture this information using expert curation, it should be indicative of evaluators' judgment.
\\\textbf{Result}:
Table~\ref{tab:vallex} shows 2 values obtained from crowdsourcing, segmented by EP ranges: 1) T(V) - the percentage of people who reported that they trust the model \emph{X} to predict valence, 2) T(Priv) - the percentage of people who trust the model \emph{X} to preserve privacy.
To answer this particular question, we analyze how evaluators choose the model that they are most likely to trust with their data for the purpose of evaluating perception of privacy preservation [T(Priv)].
The two privacy preservation methods, i.e., adversarial training and augmentation, are chosen 30\% and 26\% of the time, respectively. When asked to focus on privacy preservation the annotators choose in most of the cases.
To test the effectiveness of EP as a proxy for relative model choice, we evaluate whether the evaluators' model choices for privacy preservation change as a factor of EP. Table~\ref{tab:vallex} shows that there is a positively correlated trend between the calculated EP metric and the likelihood of a particular model being preferred. We also show how EP correlates with a random sampling of crowdsourced preferences.

\subsection{RQ3: Comparing the EP and General Metrics}
\label{sec:rq3}

\textbf{Hypothesis}: A metric that uses sociological information to integrate the degree of alignment and misalignment with known gender-biased tokens will also align with model artifacts that can be exploited to infer gender.
\\\textbf{Reason}: Unintentional capture of gender-based information in models is usually an artifact of the data used for model training~\cite{sun2019mitigating}. Given that the emotion dataset used is collected from a subset of human population, the gender biased artifacts should be captured by the proposed metric which aims to align with known sociological indicators of demographics.
\\\textbf{Result}:
We calculate EP model interpretation difference as defined in Section~\ref{sec:modelexp}. We expect GenderExp to have the lowest EP score, because it should explicitly capture gender-based information, which is should be highly correlated with the bias list described in Section~\ref{sec:iat}. We find that this model has a EP score of -.21 (Table~\ref{tab:vallex}), thus supporting the claim that our metric captures the perception of gender.
We find that augmentation methods (PrivAug) are most similar to sociolinguistic gender biases, as measured by EP.  However, we remind that both the creation of the paired augmentation lists and the resulting training time can be burdensome. The adversarial training method (PrivAdv) has the next closest EP score (relative difference of 3.97\%) but does not require the additional demographic variable replacement list annotation effort and maintains the same training time.

To show that this metric can be used as a stand-in for human judgment while still capturing human perception, we train the model (Sec~\ref{sec:genmodel}) on IEMOCAP dataset for valence generally, just optimizing for cross-entropy loss for valence prediction. For the privacy preservation model comparison, we use the adversarial variant of the model (Sec~\ref{sec:privadv}), because it relies only on the knowledge of the gender label (rather than replacement lists etc.). We incorporate our metric for privacy preservation model by two methods, (i) replacement, in which we completely remove the adversarial component of the model, and instead, use a weighted loss for maximizing both EP and Valence UAR, and, (ii) \emph{addition}, in which we retain the adversarial gender classification component, while adding an additional loss optimization for maximizing EP. 
When training and testing on IEMOCAP, the replacement method yields a valence UAR of 0.643 and gender UAR of 0.582, whereas the \emph{addition} method gives a valence UAR of 0.672 and gender UAR of 0.548. 

\subsection{RQ4: Versatility of the EP Metric}

\subsubsection{Privacy Preservation Generalization Correlation with Emotion-Privacy Metric (EP)}
\textbf{Hypothesis}: Models that have higher EP score would generalize better in their ability for privacy preservation.
\\\textbf{Reason}: Privacy preserving ability of models across datasets is has been found to be dependent on using a sample set that ideally has all associations for any demographic variable~\cite{sun2019mitigating}. Because EP is dependent on an externally curated list of such assosications by experts, rather than just relying on internally learnt associations in a black box model, it should be indicative of a model's performance on out of distribution dataset when presented with `unseen' gender-biased artificats.
\\\textbf{Results}:
In Table~\ref{tab:vallex}, we show the generalizability of the models, training on IEMOCAP or MSP-Improv+IEMOCAP (for bias fine-tuning) and testing on MuSE.  We find that the generalizability of the models with respect to both the performance of the emotion recognition (column \emph{M} Valence UAR) and efficacy of the privacy preservation (column \emph{M} Gender UAR) is affected by the method of privacy preservation.
For cross corpus performance for privacy preservation (gender UAR), we find that adversarial training paradigms are most effective for continuing to mask gender out of training distribution, compared to other privacy preserving methods.  
This is in contrast to the within-corpus setting, where data augmentation was found to be most effective.This may be because adversarial training method can learn additional replicable gender-based correlation patterns that do not agree or are unobserved in demographic studies. For future work it would be interesting to see whether these repeated patterns are dataset artifacts or new gateways into presently unknown associations.
We see a maximal drop in privacy preservation performance for PrivBias model (highest increase in attacker's ability to predict gender on MuSE use the initially trained model), which corresponds with it having the lowest \emph{EP} value amongst all privacy preservation methods.

To show that the metric is a good stand-in or supplement to privacy preservation method, we train the two models as mentioned in Section~\ref{sec:rq3} that introduce privacy either by \emph{replacement} or by \emph{addition}. When using these models trained on IEMOCAP, but testing on MuSE dataset, we obtain a valence UAR of 0.563 along with a gender UAR of 0.611, and, a valence UAR of 0.581 along with a gender UAR of 0.587 for the replacement and \emph{addition} setup respectively. We see a significantly substantial decrease in performance of the \emph{addition} setup model when predicting gender on out of distribution data, demonstrating the utility of the metric for enhanced privacy preservation while still maintaining the alignment with human perception.

\subsubsection{Emotion-Generalization (EG): Correlation with Emotion Recognition Generalization}
\textbf{Hypothesis}: EG has a positive relationship with performance of a model for cross-corpus emotion recognition.
\\\textbf{Reason}: Studies have shown that emotion recognition models have better generalization performance if they have a diverse dataset that is representative of the general emotion perception, and that the model learns these generalized patterns~\cite{kaushik2019learning}. Because EG measures the directional amount of overlap between saliency (learnt patterns) and known perception (NRC-VAD), it should be indicative of a model's generalization capability.
\\\textbf{Result}: 
As mentioned above, in Table~\ref{tab:vallex}, we see that efficacy of emotion recognition is affected when inducing privacy preservation. We find a decline in emotion recognition accuracy across corpus when inducing privacy seeing the maximal drop (10\%) in case of bias fine-tuning. 
We find a positive relationship between the probability that a model does not rely on spurious correlations in IEMOCAP, $EG$, and the cross-corpus UAR obtained on the MuSE dataset for the emotion recognition task (Table~\ref{tab:vallex}).
For example, in Table~\ref{tab:vallex}, we show that EG has the highest value for Privacy Augmentation modeling along with having the highest performance for valence prediction on MuSE to validate out of data distribution.

\section{Conclusion}
\vspace{-.7em}
We study how interpretation mechanisms can be used to evaluate privacy preserving models.
and how they can be used to inform metric design that correlates with humans' perception of model's ability for a given task. 
We analyse patterns in interpretation differences that are introduced due to an additional constraint of privacy preservation.
and look at how this difference can be directionally correlated with known sociological indicators of gender to inform our metric design. 
We conduct crowdsourcing studies to validate that this metric is representative of humans' perception of model's ability for privacy preservation. 
Finally, we show how the same metric design concept can be used for measuring generalization capability for both emotion recognition performance and privacy preservation in cross-corpus settings.


\bibliography{aaai22.bib}

\begin{thebibliography}{44}
\providecommand{\natexlab}[1]{#1}

\bibitem[{Baron and Musolesi(2020)}]{baron2020interpretable}
Baron, B.; and Musolesi, M. 2020.
\newblock Interpretable machine learning for privacy-preserving pervasive
  systems.
\newblock \emph{IEEE Pervasive Computing}, 19(1): 73--82.

\bibitem[{Belinkov and Glass(2019)}]{belinkov2019analysis}
Belinkov, Y.; and Glass, J. 2019.
\newblock Analysis methods in neural language processing: A survey.
\newblock \emph{Transactions of the Association for Computational Linguistics},
  7: 49--72.

\bibitem[{Bengio, Courville, and Vincent(2013)}]{bengio2013representation}
Bengio, Y.; Courville, A.; and Vincent, P. 2013.
\newblock Representation learning: A review and new perspectives.
\newblock \emph{IEEE transactions on pattern analysis and machine
  intelligence}.

\bibitem[{Bolukbasi et~al.(2016)Bolukbasi, Chang, Zou, Saligrama, and
  Kalai}]{bolukbasi2016man}
Bolukbasi, T.; Chang, K.-W.; Zou, J.~Y.; Saligrama, V.; and Kalai, A.~T. 2016.
\newblock Man is to computer programmer as woman is to homemaker? debiasing
  word embeddings.
\newblock In \emph{Advances in neural information processing systems}.

\bibitem[{Brescoll(2016)}]{brescoll2016leading}
Brescoll, V.~L. 2016.
\newblock Leading with their hearts? How gender stereotypes of emotion lead to
  biased evaluations of female leaders.
\newblock \emph{The Leadership Quarterly}, 27(3): 415--428.

\bibitem[{Busso et~al.(2008)Busso, Bulut, Lee, Kazemzadeh, Mower, Kim, Chang,
  Lee, and Narayanan}]{busso2008iemocap}
Busso, C.; Bulut, M.; Lee, C.-C.; Kazemzadeh, A.; Mower, E.; Kim, S.; Chang,
  J.~N.; Lee, S.; and Narayanan, S.~S. 2008.
\newblock IEMOCAP: Interactive emotional dyadic motion capture database.
\newblock \emph{Language resources and evaluation}.

\bibitem[{Busso et~al.(2016)Busso, Parthasarathy, Burmania, AbdelWahab,
  Sadoughi, and Provost}]{busso2016msp}
Busso, C.; Parthasarathy, S.; Burmania, A.; AbdelWahab, M.; Sadoughi, N.; and
  Provost, E.~M. 2016.
\newblock MSP-IMPROV: An acted corpus of dyadic interactions to study emotion
  perception.
\newblock \emph{IEEE Transactions on Affective Computing}, 8(1): 67--80.

\bibitem[{Carlini et~al.(2019)Carlini, Liu, Erlingsson, Kos, and
  Song}]{carlini2019secret}
Carlini, N.; Liu, C.; Erlingsson, {\'U}.; Kos, J.; and Song, D. 2019.
\newblock The Secret Sharer: Evaluating and testing unintended memorization in
  neural networks.
\newblock In \emph{28th $\{$USENIX$\}$ Security Symposium)}.

\bibitem[{Corbett-Davies and Goel(2018)}]{corbett2018measure}
Corbett-Davies, S.; and Goel, S. 2018.
\newblock The measure and mismeasure of fairness: A critical review of fair
  machine learning.
\newblock \emph{arXiv preprint arXiv:1808.00023}.

\bibitem[{Davidson, Bhattacharya, and Weber(2019)}]{davidson2019racial}
Davidson, T.; Bhattacharya, D.; and Weber, I. 2019.
\newblock Racial Bias in Hate Speech and Abusive Language Detection Datasets.
\newblock \emph{arXiv preprint arXiv:1905.12516}.

\bibitem[{Devlin et~al.(2018)Devlin, Chang, Lee, and
  Toutanova}]{devlin2018bert}
Devlin, J.; Chang, M.-W.; Lee, K.; and Toutanova, K. 2018.
\newblock Bert: Pre-training of deep bidirectional transformers for language
  understanding.
\newblock \emph{arXiv preprint arXiv:1810.04805}.

\bibitem[{Eckert(2017)}]{eckert2017age}
Eckert, P. 2017.
\newblock Age as a sociolinguistic variable.
\newblock \emph{The handbook of sociolinguistics}, 151--167.

\bibitem[{Engler(2021)}]{engler2021auditing}
Engler, A. 2021.
\newblock Auditing employment algorithms for discrimination.
\newblock \emph{Brookings Institute, Center for Technology Innovation}.

\bibitem[{Ferrario, Loi, and Vigan{\`o}(2019)}]{ferrario2019ai}
Ferrario, A.; Loi, M.; and Vigan{\`o}, E. 2019.
\newblock In AI we trust incrementally: a multi-layer model of trust to analyze
  human-artificial intelligence interactions.
\newblock \emph{Philosophy \& Technology}, 1--17.

\bibitem[{Fosch-Villaronga et~al.(2021)Fosch-Villaronga, Poulsen, S{\o}raa, and
  Custers}]{fosch2021little}
Fosch-Villaronga, E.; Poulsen, A.; S{\o}raa, R.~A.; and Custers, B. 2021.
\newblock A little bird told me your gender: Gender inferences in social media.
\newblock \emph{Information Processing \& Management}, 58(3): 102541.

\bibitem[{Ganin and Lempitsky(2014)}]{ganin2014unsupervised}
Ganin, Y.; and Lempitsky, V. 2014.
\newblock Unsupervised domain adaptation by backpropagation.
\newblock \emph{arXiv preprint arXiv:1409.7495}.

\bibitem[{Gerke, Minssen, and Cohen(2020)}]{gerke2020ethical}
Gerke, S.; Minssen, T.; and Cohen, I.~G. 2020.
\newblock Ethical and Legal Challenges of Artificial Intelligence-Driven Health
  Care.
\newblock \emph{Forthcoming in: Artificial Intelligence in Healthcare, 1st
  edition, Adam Bohr, Kaveh Memarzadeh (eds.)}.

\bibitem[{Heuer and Breiter(2020)}]{heuer2020more}
Heuer, H.; and Breiter, A. 2020.
\newblock More Than Accuracy: Towards Trustworthy Machine Learning Interfaces
  for Object Recognition.
\newblock In \emph{Proceedings of the 28th ACM Conference on User Modeling,
  Adaptation and Personalization}, 298--302.

\bibitem[{Iosifidis and Ntoutsi(2018)}]{iosifidis2018dealing}
Iosifidis, V.; and Ntoutsi, E. 2018.
\newblock Dealing with bias via data augmentation in supervised learning
  scenarios.
\newblock \emph{Jo Bates Paul D. Clough Robert J{\"a}schke}, 24.

\bibitem[{Jaiswal et~al.(2020)Jaiswal, Bara, Luo, Burzo, Mihalcea, and
  Provost}]{jaiswal2020muse}
Jaiswal, M.; Bara, C.-P.; Luo, Y.; Burzo, M.; Mihalcea, R.; and Provost, E.~M.
  2020.
\newblock MuSE: a Multimodal Dataset of Stressed Emotion.
\newblock In \emph{Proceedings of The 12th Language Resources and Evaluation
  Conference}, 1499--1510.

\bibitem[{Jaiswal and Provost(2020)}]{jaiswal2020privacy}
Jaiswal, M.; and Provost, E.~M. 2020.
\newblock Privacy Enhanced Multimodal Neural Representations for Emotion
  Recognition.
\newblock In \emph{AAAI}, 7985--7993.

\bibitem[{Kaushik, Hovy, and Lipton(2019)}]{kaushik2019learning}
Kaushik, D.; Hovy, E.; and Lipton, Z.~C. 2019.
\newblock Learning the difference that makes a difference with
  counterfactually-augmented data.
\newblock \emph{arXiv preprint arXiv:1909.12434}.

\bibitem[{Kim, Khanna, and Koyejo(2016)}]{kim2016examples}
Kim, B.; Khanna, R.; and Koyejo, O.~O. 2016.
\newblock Examples are not enough, learn to criticize! criticism for
  interpretability.
\newblock In \emph{Advances in Neural Information Processing Systems},
  2280--2288.

\bibitem[{Kokhlikyan et~al.(2020)Kokhlikyan, Miglani, Martin, Wang, Alsallakh,
  Reynolds, Melnikov, Kliushkina, Araya, Yan et~al.}]{kokhlikyan2020captum}
Kokhlikyan, N.; Miglani, V.; Martin, M.; Wang, E.; Alsallakh, B.; Reynolds, J.;
  Melnikov, A.; Kliushkina, N.; Araya, C.; Yan, S.; et~al. 2020.
\newblock Captum: A unified and generic model interpretability library for
  pytorch.
\newblock \emph{arXiv preprint arXiv:2009.07896}.

\bibitem[{Krishna et~al.(2020)Krishna, Khosla, Bigham, and
  Lipton}]{krishna2020generating}
Krishna, K.; Khosla, S.; Bigham, J.~P.; and Lipton, Z.~C. 2020.
\newblock Generating SOAP Notes from Doctor-Patient Conversations.
\newblock \emph{arXiv preprint arXiv:2005.01795}.

\bibitem[{Liu et~al.(2021)Liu, Ding, Shaham, Rahayu, Farokhi, and
  Lin}]{liu2021machine}
Liu, B.; Ding, M.; Shaham, S.; Rahayu, W.; Farokhi, F.; and Lin, Z. 2021.
\newblock When machine learning meets privacy: A survey and outlook.
\newblock \emph{ACM Computing Surveys (CSUR)}, 54(2): 1--36.

\bibitem[{Meng et~al.(2018)Meng, Li, Chen, Zhao, Mazalov, Gang, and
  Juang}]{meng2018speaker}
Meng, Z.; Li, J.; Chen, Z.; Zhao, Y.; Mazalov, V.; Gang, Y.; and Juang, B.-H.
  2018.
\newblock Speaker-invariant training via adversarial learning.
\newblock In \emph{2018 IEEE International Conference on Acoustics, Speech and
  Signal Processing (ICASSP)}, 5969--5973. IEEE.

\bibitem[{Mohammad(2018)}]{mohammad2018obtaining}
Mohammad, S. 2018.
\newblock Obtaining reliable human ratings of valence, arousal, and dominance
  for 20,000 english words.
\newblock In \emph{Proceedings of the 56th Annual Meeting of the Association
  for Computational Linguistics (Volume 1: Long Papers)}, 174--184.

\bibitem[{Newman et~al.(2008)Newman, Groom, Handelman, and
  Pennebaker}]{newman2008gender}
Newman, M.~L.; Groom, C.~J.; Handelman, L.~D.; and Pennebaker, J.~W. 2008.
\newblock Gender differences in language use: An analysis of 14,000 text
  samples.
\newblock \emph{Discourse Processes}, 45(3): 211--236.

\bibitem[{Nushi, Kamar, and Horvitz(2018)}]{nushi2018towards}
Nushi, B.; Kamar, E.; and Horvitz, E. 2018.
\newblock Towards accountable ai: Hybrid human-machine analyses for
  characterizing system failure.
\newblock \emph{arXiv preprint arXiv:1809.07424}.

\bibitem[{Pliev(2019)}]{emohiring}
Pliev, G. 2019.
\newblock How Emotion AI Can Transform Large-Scale Recruitment Processes.
\newblock Accessed: 10-21-2019.

\bibitem[{Preece et~al.(2018)Preece, Harborne, Braines, Tomsett, and
  Chakraborty}]{preece2018stakeholders}
Preece, A.; Harborne, D.; Braines, D.; Tomsett, R.; and Chakraborty, S. 2018.
\newblock Stakeholders in explainable AI.
\newblock \emph{arXiv preprint arXiv:1810.00184}.

\bibitem[{Pruthi et~al.(2019)Pruthi, Gupta, Dhingra, Neubig, and
  Lipton}]{pruthi2019learning}
Pruthi, D.; Gupta, M.; Dhingra, B.; Neubig, G.; and Lipton, Z.~C. 2019.
\newblock Learning to Deceive with Attention-Based Explanations.
\newblock \emph{arXiv preprint arXiv:1909.07913}.

\bibitem[{Rickford(2016)}]{rickford2016raciolinguistics}
Rickford, J.~R. 2016.
\newblock \emph{Raciolinguistics: How language shapes our ideas about race}.
\newblock Oxford University Press.

\bibitem[{Rosenberg(2012)}]{rosenberg2012classifying}
Rosenberg, A. 2012.
\newblock Classifying skewed data: Importance weighting to optimize average
  recall.
\newblock In \emph{Thirteenth Annual Conference of the International Speech
  Communication Association}.

\bibitem[{Sagawa et~al.(2020)Sagawa, Raghunathan, Koh, and
  Liang}]{sagawa2020investigation}
Sagawa, S.; Raghunathan, A.; Koh, P.~W.; and Liang, P. 2020.
\newblock An Investigation of Why Overparameterization Exacerbates Spurious
  Correlations.
\newblock \emph{arXiv preprint arXiv:2005.04345}.

\bibitem[{Schmidt and Biessmann(2019)}]{schmidt2019quantifying}
Schmidt, P.; and Biessmann, F. 2019.
\newblock Quantifying interpretability and trust in machine learning systems.
\newblock \emph{arXiv preprint arXiv:1901.08558}.

\bibitem[{Sperrle et~al.(2019)Sperrle, Schlegel, El-Assady, and
  Keim}]{sperrle2019human}
Sperrle, F.; Schlegel, U.; El-Assady, M.; and Keim, D. 2019.
\newblock Human Trust Modeling for Bias Mitigation in Artificial Intelligence.
\newblock In \emph{ACM CHI 2019 Workshop: Where is the Human? Bridging the Gap
  Between AI and HCI}.

\bibitem[{Sun et~al.(2019)Sun, Gaut, Tang, Huang, ElSherief, Zhao, Mirza,
  Belding, Chang, and Wang}]{sun2019mitigating}
Sun, T.; Gaut, A.; Tang, S.; Huang, Y.; ElSherief, M.; Zhao, J.; Mirza, D.;
  Belding, E.; Chang, K.-W.; and Wang, W.~Y. 2019.
\newblock Mitigating gender bias in natural language processing: Literature
  review.
\newblock \emph{arXiv preprint arXiv:1906.08976}.

\bibitem[{Sundararajan, Taly, and Yan(2017)}]{sundararajan2017axiomatic}
Sundararajan, M.; Taly, A.; and Yan, Q. 2017.
\newblock Axiomatic attribution for deep networks.
\newblock \emph{arXiv preprint arXiv:1703.01365}.

\bibitem[{Swinger et~al.(2019)Swinger, De-Arteaga, Heffernan~IV, Leiserson, and
  Kalai}]{swinger2019biases}
Swinger, N.; De-Arteaga, M.; Heffernan~IV, N.~T.; Leiserson, M.~D.; and Kalai,
  A.~T. 2019.
\newblock What are the biases in my word embedding?
\newblock In \emph{Proceedings of the 2019 AAAI/ACM Conference on AI, Ethics,
  and Society}, 305--311.

\bibitem[{Szegedy et~al.(2013)Szegedy, Zaremba, Sutskever, Bruna, Erhan,
  Goodfellow, and Fergus}]{szegedy2013intriguing}
Szegedy, C.; Zaremba, W.; Sutskever, I.; Bruna, J.; Erhan, D.; Goodfellow, I.;
  and Fergus, R. 2013.
\newblock Intriguing properties of neural networks.
\newblock \emph{arXiv preprint arXiv:1312.6199}.

\bibitem[{Wang and Chang(2020)}]{wang2020privacy}
Wang, S.; and Chang, J.~M. 2020.
\newblock Privacy-Preserving Image Classification in the Local Setting.
\newblock \emph{arXiv preprint arXiv:2002.03261}.

\bibitem[{Wolf et~al.(2019)Wolf, Debut, Sanh, Chaumond, Delangue, Moi, Cistac,
  Rault, Louf, Funtowicz et~al.}]{wolf2019huggingface}
Wolf, T.; Debut, L.; Sanh, V.; Chaumond, J.; Delangue, C.; Moi, A.; Cistac, P.;
  Rault, T.; Louf, R.; Funtowicz, M.; et~al. 2019.
\newblock HuggingFace's Transformers: State-of-the-art Natural Language
  Processing.
\newblock \emph{ArXiv}, arXiv--1910.

\end{thebibliography}

\end{document}